\documentclass[runningheads]{llncs}
\usepackage{graphicx}
\usepackage[caption=false]{subfig}
\usepackage[utf8]{inputenc}
\usepackage{amsfonts}
\usepackage{amsmath}
\usepackage{siunitx}

\newcommand{\R}{\mathbb{R}}

\newcommand{\new}[1]{#1}

\newcommand{\norm}[1]{\left\lVert#1\right\rVert}

\newcommand{\mat}[1]{\mathrm{#1}}
\newcommand{\vett}[1]{\boldsymbol{\mathbf{#1}}}
\newcommand{\qot}[1]{``#1''}

\newcommand{\dotp}[2]{\langle #1, #2 \rangle}

\DeclareMathOperator*{\argmin}{arg\,min}

\begin{document}
\title{PRO-TIP: Phantom for RObust automatic ultrasound calibration by TIP detection}
\titlerunning{PRO-TIP: Automatic Ultrasound Calibration}
\author{Matteo Ronchetti\inst{1} \and
Julia Rackerseder \inst{1} \and Maria Tirindelli\inst{1,2} \and Mehrdad Salehi\inst{2} \and Nassir Navab\inst{2} \and Wolfgang Wein\inst{1}  \and Oliver Zettinig \inst{1}}

\authorrunning{M. Ronchetti et al.}
\institute{ImFusion GmbH, M\"unchen, Germany
\and Computer Aided Medical Procedures (CAMP) Technische Universit\"at M\"unchen, Germany}
\maketitle              %

\begin{abstract}
We propose a novel method to automatically calibrate tracked ultrasound probes.
To this end we design a custom phantom consisting of nine cones with different heights.
The tips are used as key points to be matched between multiple sweeps.
We extract them using a convolutional neural network to segment the cones in every ultrasound frame and then track them across the sweep.
The calibration is robustly estimated using RANSAC and later refined employing image based techniques.
Our phantom can be 3D-printed and offers many advantages over state-of-the-art methods. 
The phantom design and algorithm code are freely available online.
Since our phantom does not require a tracking target on itself, ease of use is improved over currently used techniques.
The fully automatic method generalizes to new probes and different vendors, as shown in our experiments.
Our approach produces results comparable to calibrations obtained by a domain expert.

\end{abstract}

\section{Introduction}
Ultrasound (US) imaging is a widely used medical imaging modality.
Due to its real-time, non-radiation-based imaging capabilities, it gains ever more popularity in the medical domain.
By combining an ultrasound transducer with a tracking system, 3D free-hand acquisitions become possible, further increasing the range of possible applications, for instance intra-operative navigation.
A tracking system consists of a tracking device, %
which is able to monitor the position of a tracking target (\qot{marker}) in its own coordinate system~\cite{chen2020external}.
The marker is rigidly attached to a US probe, but regularly in an arbitrary fashion relative to the image coordinate system.
The geometrical relation mapping from the image space into the marker's local coordinate system is commonly referred to as (spatial) ultrasound calibration.

\new{Accurate calibration often turns out to be a tedious task,
it is therefore not surprising that numerous approaches have been devised in the past to solve this problem~\cite{mercier2005review}.} 
Many require additional tools or objects, which may also be tracked themselves.
The simplest approach requires a tracked and pivot-calibrated stylus, which is held such that its tip is visible in the US image at multiple locations~\cite{muratore2001beam}.
This technique has been improved by also digitizing points on the probe itself~\cite{viswanathan2004immediate}, or by enforcing stylus orientation through a phantom~\cite{wen2020novel}.
A majority of works utilize dedicated phantoms, i.e. objects with known geometric properties, for achieving the calibration~\cite{hsu2009freehand}.
\new{Such phantoms include sphere-like objects~\cite{barratt2001accuracy}, pairs of crossing wires~\cite{detmer19943d}, multiple point targets~\cite{meairs2000reconstruction}, planes~\cite{dandekar2005phantom}, LEGO bricks~\cite{xiao2016user}, wires~\cite{wire_phantom,wire_phantom_new,carbajal2013improving,shen2019method} and phantoms actively transmitting US echo back to the transducer~\cite{cheng2017active}.}
Using crossing wires as a phantom, automatic robotic calibration using visual servoing has also been demonstrated~\cite{krupa2006automatic}.
Unfortunately, the described auxiliary control task is often not implementable on medical robotic systems as the required low-level joint control is encapsulated by the vendor.
The aforementioned methods may be characterized by one or multiple disadvantages.
First, manufacturing tolerances of phantoms and styluses (or the technique to calibrate them), as well as relative tracking between two tracking targets give rise to calibration errors.
Phantoms may be tricky to assemble and geometrically characterize in code, for instances using the wires lead through the holes of 3D-printed objects.
Finally, some approaches limit possible probe orientations, e.g. due to space between the wire phantom walls, which results in under-sampling of the six degrees of freedom, in turn leading to different in-plane and out-of-plane accuracies.
In contrast to tool- or phantom-based calibration, image-based techniques work with arbitrary objects~\cite{image_based_calibration}.
The idea resembles image registration as the similarity of US images acquired from different directions is iteratively maximized.
This method has been shown to achieve high calibration accuracies for static objects but, just like registration problems, suffers from a limited capture range and hence needs to be initialized well.

\begin{figure}[t]
\centering
\subfloat[]{
    \label{fig:phantom}
   \includegraphics[width=0.45\textwidth]{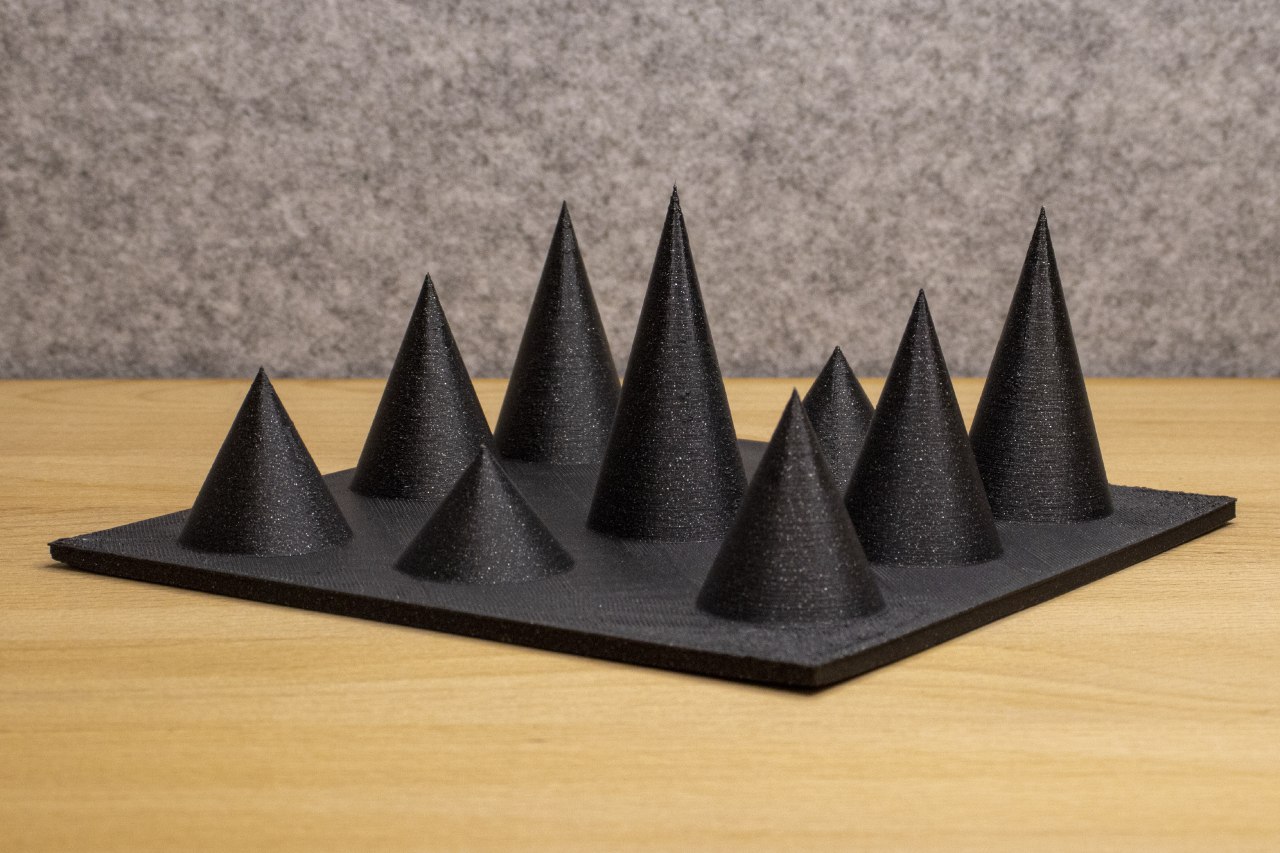}
}
\subfloat[]{
    \label{fig:labelmap_3d}
   \includegraphics[width=0.45\textwidth]{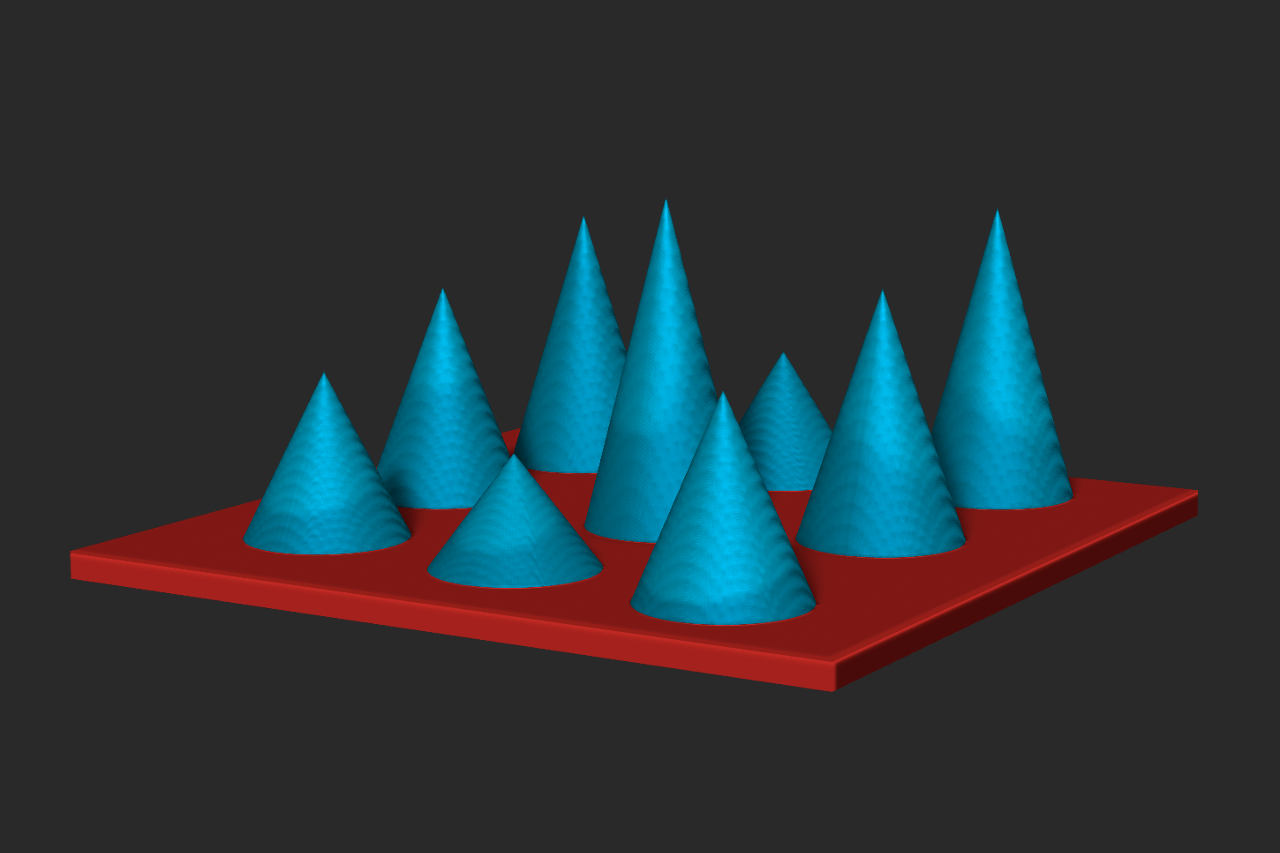}
}
\vspace{-0.75\baselineskip}
\caption{A photograph of the proposed 3D-printed phantom (a) and its corresponding labelmap (b), showing cones in blue and the base plate in red color.}
\end{figure}
\new{In this work, we propose a novel calibration technique that uses both feature- and image-based calibration on the same 3D-printed phantom.}
The phantom consists of nine cones and is compatible with a wide range of transducer designs and imaging depths.
\new{A machine learning model detects the cone tips in two sets of recorded US images. These are then matched and used to estimate a rigid calibration matrix, which is subsequently refined with an image-based method.
The proposed method does not require any additional tracked tool or tracking on the phantom and no assembly other than 3D-printing, therefore it is faster and easier to use than existing approaches. 
In contrast to state of the art feature-based approaches, we use image-based refinement, which does not require tight manufacturing tolerances on the phantom.
Furthermore, the usage of feature-based calibration as initialization for an image-based method removes the necessity of manual initialization, making the process fully automatic. \\
The CAD model for 3D-printing our phantom, the machine learning model for segmentation, as well as a reference implementation are freely available online\footnote[1]{https://github.com/ImFusionGmbH/PRO-TIP-Automatic-Ultrasound-Calibration}}

\section{Approach}
The main idea behind the proposed method is to automatically detect multiple distinguishable keypoints in two sweeps, i.e. freehand US recordings slowly sweeping over the phantom, and match them to estimate the calibration matrix.

\subsection{Phantom and Model Preparation}
\label{sec:preparation}

\noindent\textbf{Phantom design}
We design a phantom composed of nine cones of different heights as depicted in Fig.~\ref{fig:phantom}.
The tips of the cones are used as keypoints, uniquely identifiable by their height from the base plane.
To avoid confusion of cones of similar size, we make sure tips with comparable heights are not placed next to each other.
\new{Cone heights are spaced uniformly in the range $[20mm, 60mm]$, a wider range would make cones more distinguishable but would also limit the range of usable imaging depths. We have found that using 9 cones is a good compromise between the number of keypoints and the number of mismatches due to similar cone heights. }
Our phantom design can be easily 3D-printed and produced in different scales depending on the image shape and depth.

\noindent\textbf{Model architecture and training}
A convolutional neural network \cite{cnn,cnn_2,cnn_3} is used to segment every frame of the sweeps, distinguishing background, cones and base plane.
We follow the U-Net \cite{unet} architecture and make use of residual blocks \cite{resnet}, leaky ReLUs \cite{leaky_relu} and instance normalization \cite{instance_normalization}.
The network does not make use of the temporal contiguity of frames but processes each frame independently.
Augmentation on the training data is used to make the model as robust as possible against reflections and changes of probe.
\new{In particular, we re-scale every frame to $1mm$ spacing, crop a random $128\times128$ section, apply Cutout~\cite{cutout} with random intensity, add speckle noise at a resolution of $64\times32$ and Gaussian noise at full resolution.}

\noindent\textbf{Data labelling}
We manually register tracked sweeps to the phantom's 3D label map (Fig.~\ref{fig:labelmap_3d}) and use volume re-slicing to obtain 2D segmentation maps for every frame,
\new{which} allows us to quickly obtain a large dataset of labelled frames.

\begin{figure}[t]
\centering

\subfloat[]{
    \label{fig:frame_detection}
   \includegraphics[width=0.5\textwidth]{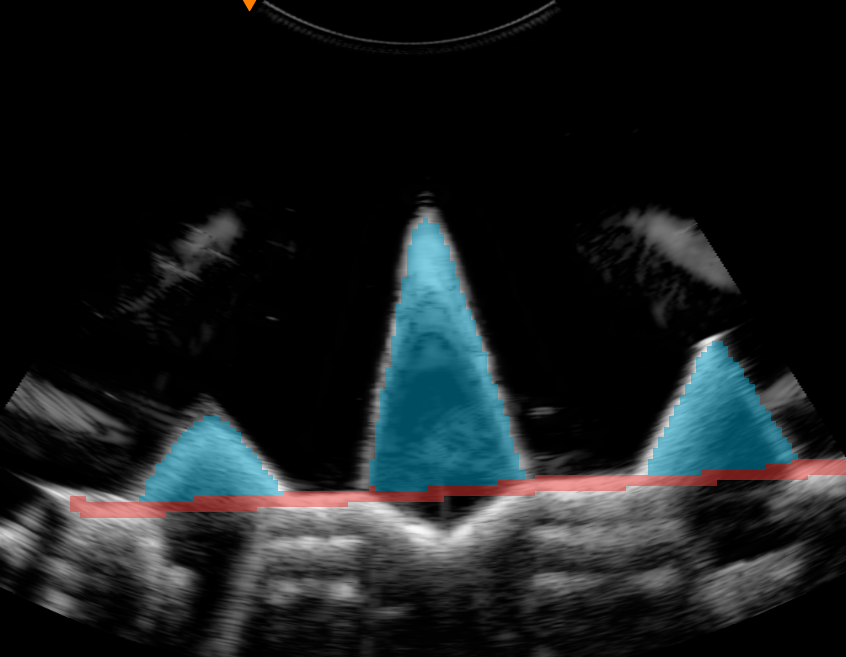}
}
\subfloat[]{
    \label{fig:reconstruction}
   \includegraphics[width=0.5\textwidth]{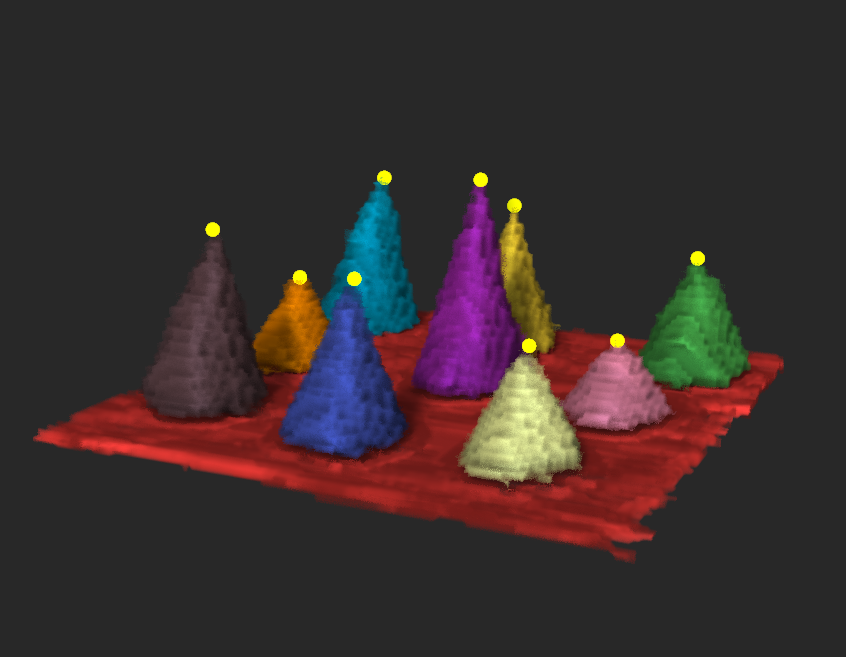}
}
\vspace{-0.75\baselineskip}
\caption{Output of the segmentation model (a) and a compounded 3D labelmap with tip detections (b). Every \textit{track} in (b) is visualized with a different color, distinguishing the different cones well. \textit{See text for details.}}
\end{figure}

\subsection{Calibration Method}
The required input for the proposed method consists of two sets of freehand ultrasound frames, denoted $A$ and $B$, from different principal directions (e.g. axial and sagittal), each covering ideally all phantom cones.
The phantom must remain static throughout all recordings.
We assume the image scale to be known and only consider a rigid calibration.

\noindent\textbf{Tip detection and matching}
After segmentation, every frame is processed to detect and track cones.
First, a line is fitted to the base plane using RANSAC \cite{ransac}, if the number of inliers is too low the frame is skipped.
Pixels located underneath the base and  small connected components ($<25mm^2$) are discarded.
The remaining connected components are considered cone detections and are tracked from frame to frame.
We denote a list of detections corresponding to the same cone a \textit{track}.
Assignment of detections to tracks is done based on the area of the intersection of the bounding boxes. A new track is created every time a detection does not intersect with any existing track. Tracks are terminated if they do not receive new detections for more than 3 frames, as shown in Fig.~\ref{fig:reconstruction}.
After the last frame has been processed, tracks with less than 10 detections are discarded.
For each detection in a track, we store the position of the highest point and its associated height. 
A mean smoothing filter is used to reduce the noise on height measurements.
The point with maximum measured height is considered the tip of the cone track.
Tips shorter than $1.5cm$ or localized outside of the US geometry (when using convex or sector probes) are discarded.
Tips from different sweeps are considered matches if their measured height difference is less than a threshold of $3mm$.

\noindent\textbf{Calibration estimation from tip matches}
Given a point $\vett{p} = (x, y, 0, 1)^T$ on the i-th frame of an ultrasound sweep, its real world position is $\vett{q} = \mat{T}_i \mat{C} \vett{p} \in \R^4$, where $\mat{T}_i$ is the tracking matrix (mapping from the tracking target to the world origin) and $C$ is the calibration matrix.
Our goal is to estimate the calibration matrix $C$ given $n$ pairs of corresponding tip detections on frames.
\\
Let $\vett{p}_i^A = (x_i^A, y_i^A, 0, 1)$ be the i-th tip detection on sweep A and $p_i^B$ be the corresponding detection on the second sweep, these points should correspond to the same world position.
Therefore we seek to minimize $\sum_i \norm{\mat{T}_i^A \mat{C} \vett{p}_i^A - \mat{T}_i^B \mat{C} \vett{p}_i^B}^2$ --- a linear least-squares problem constrained by $C$ being a rigid transformation.
\\
We manipulate the problem in order to bring it into the canonical least-squares formulation.
The tracking and calibration matrices can be divided into blocks
$$
\mat{T} = \left( \begin{array}{c|c}
   \mat{\Phi} & \vett{\delta}
   \\ \hline 0 & 1
 \end{array} \right)
 \qquad
\mat{C} = \left( \begin{array}{c|c}
   \mat{R} & \vett{t} \\ \hline 0 & 1
 \end{array} \right) \qquad \mat{R} = \left( \begin{array}{ccc}
| & | & | \\ 
\vett{u} & \vett{v} & \vett{w} \\
| & | & | \\
 \end{array} \right)
$$
where $\mat{\Phi}, \mat{R} \in \R^{3\times 3}$ are rotation matrices, and $\vett{t}, \vett{\delta} \in \R^3$ are translation vectors.
Note that, because $\mat{R}^T \mat{R} = I$ and $\text{det}(\mat{R}) = 1$, the column $\vett{w}$ must be $\vett{w} = \vett{u} \times \vett{v}$.
Then:
\begin{gather*}
\sum_i \norm{\mat{T}_i^A \mat{C} \vett{p}_i^A - \mat{T}_i^B \mat{C} \vett{p}_i^B}_2^2 = \\
= \sum_i \norm{(x^A_i \mat{\Phi}^A_i - x^B_i \mat{\Phi}^B_i) \vett{u} + (y^A_i \mat{\Phi}^A_i - y^B_i \mat{\Phi}^B_i) \vett{v} + (\mat{\Phi}^A_i - \mat{\Phi}^B_i) \vett{t} + \vett{\delta}^A_i - \vett{\delta}^B_i}_2^2\\
= \norm{
 \underbrace{\left( \begin{array}{c|c|c}
x^A_1 \mat{\Phi}^A_1 - x^B_1 \mat{\Phi}^B_1 & y^A_1 \mat{\Phi}^A_1 - y^B_1 \mat{\Phi}^B_1 & \mat{\Phi}^A_1 - \mat{\Phi}^B_1 \\
\vdots & \vdots & \vdots \\
x^A_n \mat{\Phi}^A_n - x^B_n \mat{\Phi}^B_n & y^A_n \mat{\Phi}^A_n - y^B_n \mat{\Phi}^B_n & \mat{\Phi}^A_n - \mat{\Phi}^B_n \\
 \end{array} \right)}_{=\mat{A}}
\left( \begin{array}{c}
\vett{u} \\ \vett{v} \\ \vett{t} \end{array}\right) -  \underbrace{\left( \begin{array}{c}
\vett{\delta}^B_1 - \vett{\delta}^A_1 \\ \vdots \\ \vett{\delta}^B_n - \vett{\delta}^A_n \end{array}\right)}_{=\vett{d}}}_F^2
\end{gather*}
We can then write the calibration problem as a constrained linear least squares problem with matrix of size $3n \times 9$:
\begin{align}
\label{eq:problem}
\argmin_{\vett{u}, \vett{v}, \vett{t}} \norm{\mat{A} \left( \begin{array}{c}
\vett{u} \\ \vett{v} \\ \vett{t} \end{array}\right) - \vett{d}}_F^2 \quad \text{subject to} \quad \norm{\vett{u}} = \norm{\vett{v}} = 1 \text{ and } \dotp{\vett{u}}{\vett{v}} = 0 \,.
\end{align}
We solve the unconstrained problem, and project the solution $(\vett{u}', \vett{v}', \vett{t}')$ onto the constraints using SVD \cite{matrix_nearness} obtaining $\vett{u}^*$ and $\vett{v}^*$.
The translation $\vett{t}^*$ is then computed by solving the least squares problem while keeping $\vett{u}=\vett{u}^*$ and $\vett{v}=\vett{v}^*$ fixed.
While it would be possible to iteratively improve the solution $(\vett{u}^*, \vett{v}^*, \vett{t}^*)$, for example using projected gradient, our approximation delivers results of sufficient quality for the image-based refinement to converge.

\noindent\textbf{RANSAC estimation}
The tip matching procedure is subject to errors, therefore we use RANSAC \cite{ransac} to robustly estimate the calibration matrix, disregarding the effect of erroneous matches.
At every RANSAC iteration, we sample 4 pairs of tips and use them to produce a calibration hypothesis as described in the previous section.
Although 3 pairs would be enough to solve Eq.~\ref{eq:problem}, we observed that, because of noise and bad conditioning of the problem, it is beneficial to use 4 pairs.
If the residual error of Eq.~\ref{eq:problem} is more than 100, the calibration hypothesis is immediately discarded.
Otherwise the score of the hypothesis is computed by counting the numbers of inliers, i.e. pairs such that $\norm{\mat{T}_i^A \mat{C} \vett{p}_i^A - \mat{T}_i^B \mat{C} \vett{p}_i^B}^2 \leq 1cm$.
The hypothesis with the highest score is refined by solving Eq.~\ref{eq:problem} again using all the inliers.

\noindent\textbf{Image based refinement}
The structure of our phantom produces ultrasound frames that exhibit many clear structures, therefore the calibration can be improved by using an image-based method.
In particular, we use the method described in \cite{image_based_calibration} on the same sweeps used to compute the calibration.
This method refines the initial calibration to maximize the 2D normalized cross-correlation between frames of one sweep and in-plane reconstructions of all the slices of the other sweep.

\noindent\textbf{Implementation details}
The algorithm is implemented as plugin to the ImFusion Suite 2.36.3 (ImFusion GmbH, Germany) using their SDK, including off-the-shelf implementations of the image-based calibration, volume reslicing, and PyTorch 1.8.0 integration for the machine learning model.

\noindent\textbf{Limitations}
\new{
Our algorithm needs the cone's tip and base to be visible in the same frame, in order to estimate its height. This limits the range of angles that the probe can cover and might limit the accuracy of the produced calibration.
Our method is composed of multiple components which increase complexity of implementation. We mitigate this issue by releasing our source code.
The proposed phantom design is not completely generic and would need to be adapted to specialized transducers for example end- or side-fire probes.}

\section{Experiments and Results}
\label{sec:experiments}

\begin{figure}[t]
\centering
\subfloat[]{
\includegraphics[width=0.20\textwidth]{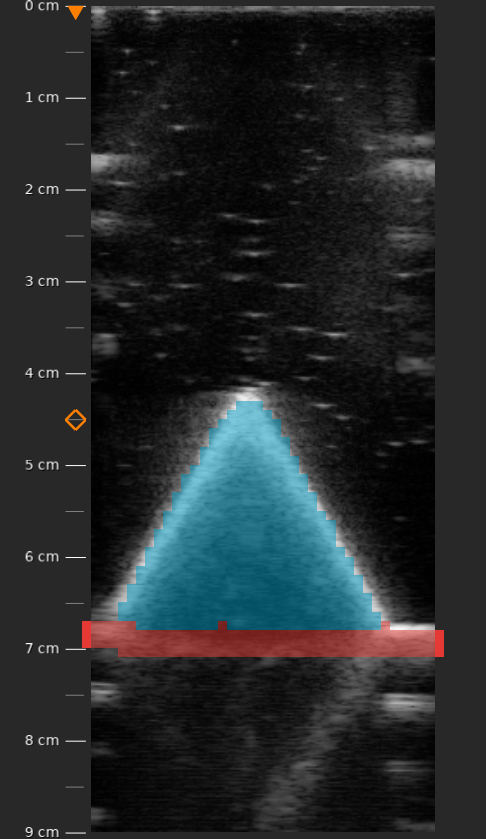}
}
\subfloat[]{
\includegraphics[width=0.406\textwidth]{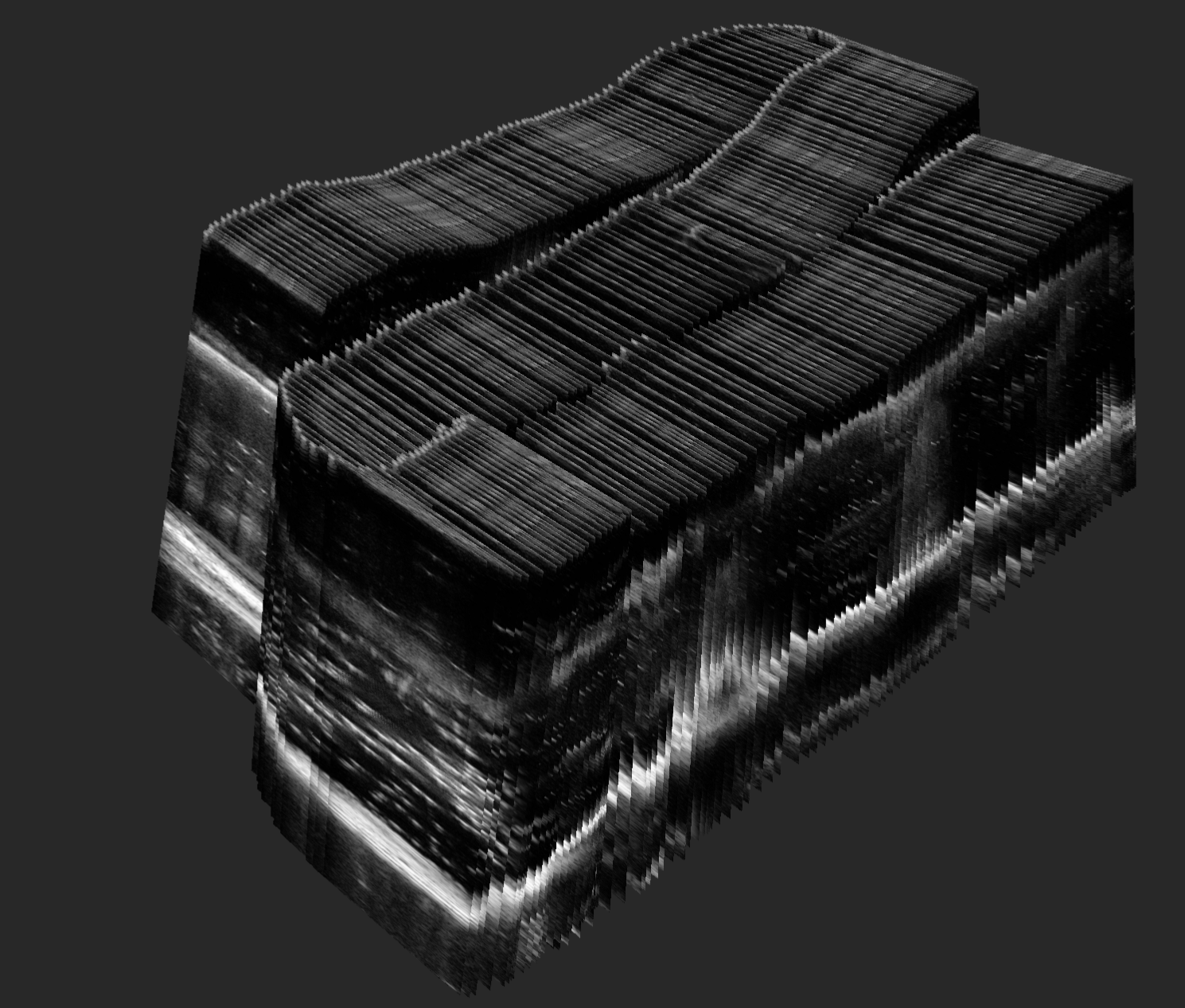}
}
\caption{A frame taken with the Clarius linear probe segmented by our model (a) and a 3D rendering of one of the sweeps used for the calibration of the Clarius probe (b). The image (a) exhibits different shape and content from frames taken with a convex probe (Figure 
\ref{fig:frame_detection}).
The sweeping motion (b) required to cover the whole phantom is unique to the linear probe and shows the generalization capabilities of our algorithm. }
\label{fig:clarius_linear}
\end{figure}

\begin{figure}[t]
\centering
\includegraphics[width=0.65\textwidth]{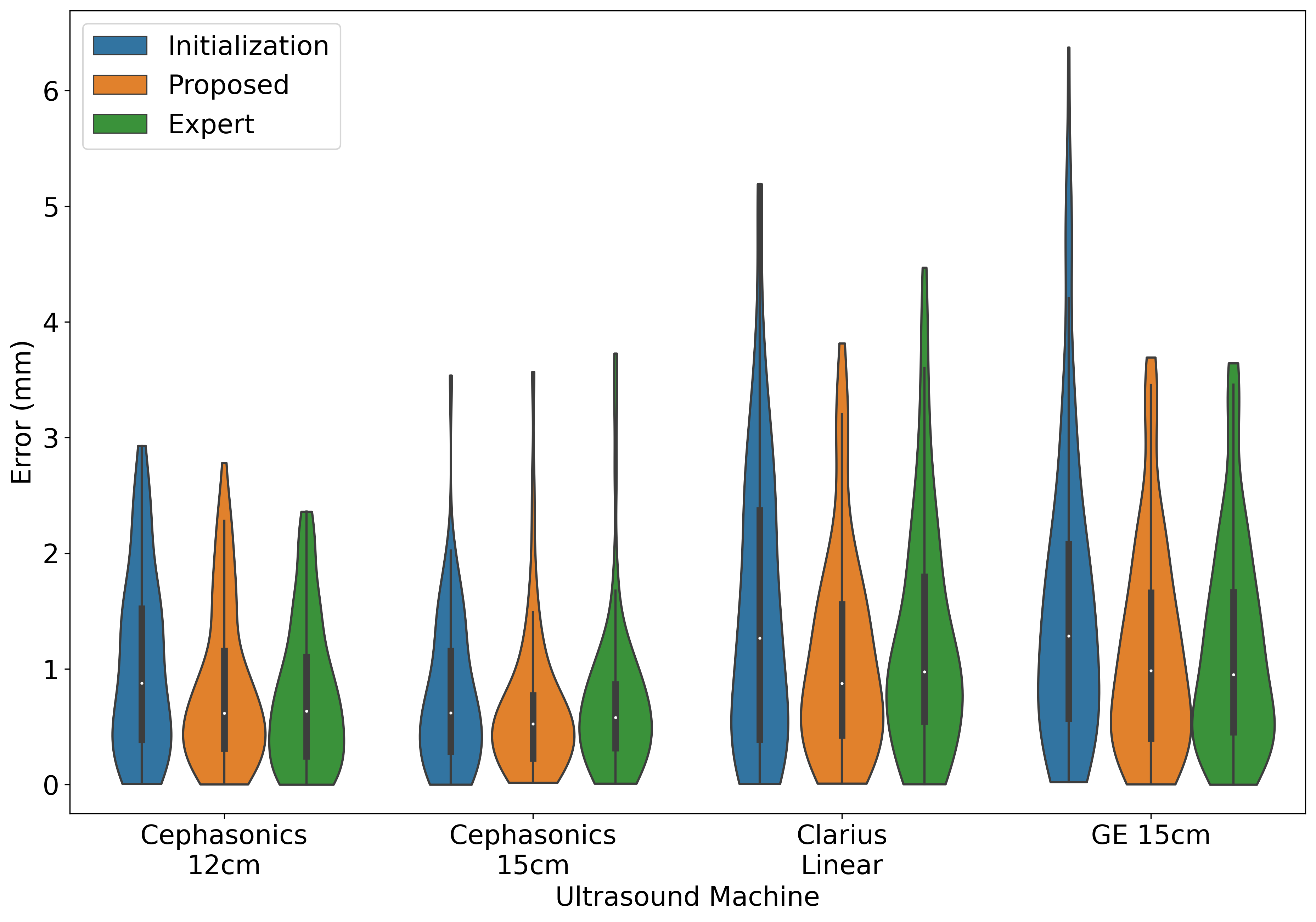}
\vspace{-0.75\baselineskip}
\caption{
Error distributions of fiducial pair distances of our method before (blue) and after image-based refinement (orange), and expert calibration with refinement (green) on data collected with different US systems.
}
\label{fig:results}
\end{figure}

\begin{table}[t]
\centering
\renewcommand\arraystretch{1.25}
\setlength{\tabcolsep}{1.5mm}
\begin{tabular}{l|c|cc|c}
& Initialization & \multicolumn{2}{c|}{Proposed} & Expert\\
& Error & Error & Time & Error \\
\hline
Cephasonics 12cm & 0.88mm & 0.64mm & 17.3s & \textbf{0.63mm} \\
Cephasonics 15cm & 0.62mm & \textbf{0.52mm} & 15.9s & 0.58mm \\
Clarius Linear &   1.27mm & \textbf{0.87mm} & 14.0s &0.96mm\\
GE 15cm &  1.28mm  & 0.98mm & 24.4s & \textbf{0.95mm} \\
\end{tabular}
\vspace{0.5\baselineskip}
\caption{Median fiducial pair distances of our method before and after image-based refinement, compared to the expert calibration.}
\label{table_results}
\end{table}

\noindent\textbf{Dataset}
We collect data using a variety of probes and tracking systems, to increase robustness of our method.
Ultrasound frames are labelled as described in Section \ref{sec:preparation}. Frames that do not show the base of the phantom or cones are discarded.
We use the following hardware combinations:
(1) Cicada research system (Cephasonics Inc., CA, USA) with convex probe tracked using a proprietary optical tracking system (Stryker Leibinger GmbH \& Co KG, Germany).
(2) LOGIQ E10S commercial system (GE Healthcare Inc., IL, USA) with a convex abdominal probe (GE C1-6) tracked using an Universal Robots, Denmark, UR5e robotic arm.
(3) Clarius L7HD handheld portable linear probe (Clarius Mobile Health Inc., BC, Canada) and (4) Clarius C3 HD3 convex probe, both tracked using the same camera as in (1).
\new{We use sweeps from (1) and (2) as training data (2492 frames) except for two sweeps from each probe that are used for validation and testing (2095 frames). Sweeps from (3) and (4) are used exclusively for testing the entire calibration pipeline.}
By giving the same input sweeps as our method to a domain expert, we obtain a reference calibration.
The expert manually initializes the calibration parameters, which consequently are optimized with the imaged-based calibration. 
These steps are potentially repeated until satisfactory overlap of sweep contents is achieved. We refer to the results as \qot{Expert} calibration.
For the evaluation, the cone tips are manually annotated in all sweeps by two users, and their positions after calibration in world coordinates are used as fiducials.

\noindent\textbf{Results}
We first evaluate whether the detected cone tips are correctly identified.
The distance between cone tip annotations and detected tips are on average 0.72 frames apart
\new{A few wrongly detected cones (later removed by RANSAC) cause the average distance to be $10.3$mm, but the median distance of $1.45$mm is well within the capture range of image-based refinement.}
The aim of our method is not to achieve better accuracy than current state-of-the-art methods, but rather match the results achieved by experienced users in a fraction of the time and without any human intervention.
Therefore, we compare our results to the \qot{Expert} calibration.
Our calibration method is evaluated both with (\qot{Proposed}) and without the image-based refinement (\qot{Initialization}).
We compute the Euclidean distances between all ${{9}\choose{2}}=36$ possible fiducial pairs and compare these to the ground truth distances of the phantom.
Figure~\ref{fig:results} depicts the distribution of errors obtained by different approaches on the datasets.
It can be noticed that the results obtained by the proposed method achieve similar results to the expert calibration, both in average error and distribution shape.
The comparably good results without image-based refinement show that our method can reliably produce an initialization within the narrow basin of attraction of the optimization method.
Of particular interest are the results with the Clarius probe data, because frame geometry and sweep motion are different from any data used as training input (Fig.~\ref{fig:clarius_linear}).
Table \ref{table_results} shows numerical results together with execution time of the Proposed method measured on a laptop running an Nvidia GTX1650. 
Runtime of our fully automatic method is in the order of seconds, while the manual approach takes several minutes.

\section{Conclusion}
In this paper, we proposed a fully automatic yet accurate approach to ultrasound calibration.
We showed that our results are comparable to calibrations done by domain experts.
Furthermore, the novel phantom design is easy to manufacture and use.
This makes calibration accessible to a broader audience and keeps the overall time spent on this task low.
Potential future extensions include phantom design adaptations to more specialized transducers, to simultaneously estimate also the temporal calibration between image and tracking source, and corrections of  speed-of-sound.\\

\noindent\textbf{Acknowledgment}
This work was partially funded by the German Federal Ministry of Education and Research (BMBF),
grant 13GW0293B (\qot{FOMIPU}).

\newpage

\bibliographystyle{splncs04}
\bibliography{citations}

\end{document}